\documentclass{article}

\usepackage{PRIMEarxiv}

\usepackage[utf8]{inputenc} 
\usepackage[T1]{fontenc}    
\usepackage{hyperref}       
\usepackage{url}            
\usepackage{booktabs}       
\usepackage{amsfonts}       
\usepackage{nicefrac}       
\usepackage{microtype}      
\usepackage{lipsum}
\usepackage{fancyhdr}       
\usepackage{graphicx}       
\usepackage{natbib}
\usepackage{tikz}
\usepackage{doi}
\usepackage{algorithm}
\usepackage{algpseudocode}
\usepackage{listings}
\graphicspath{{media/}}     

\newtheorem{theorem}{Theorem}
\newtheorem{definition}{Definition}
\newtheorem{proposition}{Proposition}

\newcommand{\Args}{\mathcal{A}}
\newcommand{\Contexts}{\mathcal{C}}

\title{From Contexts to Values: Context-Dependent Defeat in Abstract Argumentation}

\author{
  Albert Sadowski \\
  Faculty of Electronics and Information Technology \\
  Warsaw University of Technology \\
  Warsaw, Poland \\
  \texttt{albert.sadowski.stud@pw.edu.pl} \\
   \And
  Jarosław A. Chudziak \\
  Faculty of Electronics and Information Technology \\
  Warsaw University of Technology \\
  Warsaw, Poland \\
  \texttt{jaroslaw.chudziak@pw.edu.pl} \\
}

\begin{document}
\maketitle

\begin{abstract}
In value-based argumentation, an audience's ordering of values decides which attacks succeed as defeats. In many settings the deciding factor is not the audience but the circumstances: the same attack may succeed at one procedural stage, or under one regulation, and fail at another. Context-dependent argumentation frameworks (CDAFs), a model we recently introduced, capture this directly: one attack relation and a defeat function that switches each attack on or off per context, so every context induces an ordinary Dung framework. This raises a reduction question: can one value assignment with per-context orderings reproduce the defeat function, collapsing the CDAF into a VAF? We present a polynomial-time decision procedure for this question and map the harder neighbouring problems, with upper bounds from NP to $\Sigma^p_3$. We also present a validated reference implementation and a measurement: representability is rare and falls fast with the number of contexts.
\end{abstract}

\keywords{abstract argumentation \and value-based argumentation \and computational argumentation}

\section{Introduction}\label{sec:intro}

In abstract argumentation, a Dung framework fixes a set of arguments and an attack relation, and acceptance is computed from that fixed structure \cite{dung1995}. Value-based argumentation frameworks (VAFs) added an influential source of variation: arguments promote values, an audience orders the values, and the ordering decides which attacks succeed as defeats \cite{benchCapon2003}. One structure, many audiences, many outcomes.

In many settings the variation does not come from who evaluates, but from the circumstances under which the evaluation happens. A safety argument may defeat a cost argument during an incident and fail to defeat it during routine planning. The arguments and the attack stay the same; what changes is whether the attack succeeds. The circumstance is external to the arguments, and existing formalisms do not take it as a parameter. In recent work we introduced context-dependent argumentation frameworks (CDAFs) to model it directly \cite{sadowski2026choosinglensstrategicperspective}: a defeat function $\delta$ marks each attack as operative or suppressed in each context, so a CDAF is a finite family of Dung frameworks over one shared attack structure.

The family view raises three questions. First, is contextual variation genuinely new? VAFs already vary defeat through audiences, so a CDAF might be a VAF in disguise: one value assignment and one ordering per context reproducing $\delta$ exactly. When does this reduction hold, and how fast can it be decided? Second, when $\delta$ cannot be reproduced, can a VAF at least reproduce the extensions of every induced framework, and at what cost? Third, from a systems perspective: can the decision procedure be implemented and validated, and how common is reducibility?

Context appears in argumentation in several roles: as modules that constrain one another \cite{brewka2009}, or as agents' backgrounds moving toward consensus \cite{Yu2023}. Closer to our question, other extensions vary which attacks succeed, but internally or epistemically: extended frameworks argue about preferences inside the framework \cite{modgil2009}, and incomplete frameworks induce a family of completions read as possible worlds \cite{Baumeister2021}. Property-based preferences derive an argument ordering from properties attached to arguments \cite{booth2013property}; a value assignment is the special case of one property per argument, and our value-consistent partition fixes the granularity at which such a derived ordering can separate arguments. A CDAF's variation is external: contexts are a parameter, and the family is evaluated as a whole.

Section \ref{sec:representability} gives a structural criterion, a decision procedure running in $O(n^2 k + nmk)$ time for $n$ arguments, $m$ attacks, and $k$ contexts, and a trace on an instance built to exercise every phase. Section \ref{sec:complexity} maps the realisability landscape, where the upper bounds run from NP to $\Sigma^p_3$ with no matching lower bounds. Section \ref{sec:benchmarks} presents a validated reference implementation and a measurement. Section \ref{sec:open} lists open problems.

\section{Preliminaries}\label{sec:prelim}

An \emph{argumentation framework} (AF) is a pair $F = \langle \Args, R \rangle$ with $\Args$ a finite set of arguments and $R \subseteq \Args \times \Args$ an attack relation \cite{dung1995}. A semantics $\sigma$ assigns to $F$ a set $\sigma(F)$ of \emph{extensions}, subsets of $\Args$. We use the grounded, complete, preferred, and stable semantics with their standard definitions, and refer to \citet{baroni2018}. Throughout, $n = |\Args|$ and $m = |R|$.

A \emph{value-based argumentation framework} (VAF) extends an AF with a finite set of values $V$, an assignment $\mathit{val}\colon \Args \to V$, and a set of audiences \cite{benchCapon2003}, each audience $p$ being a strict total order $\succ_p$ on $V$. An attack $(a,b) \in R$ succeeds as a \emph{defeat} for $p$ iff $\mathit{val}(b) \not\succ_p \mathit{val}(a)$, so it fails only if the attacked argument carries a strictly preferred value. Audience $p$ thus induces the defeat relation $R_p = \{(a,b) \in R : \mathit{val}(b) \not\succ_p \mathit{val}(a)\}$.

\begin{definition}[\cite{sadowski2026choosinglensstrategicperspective}]\label{def:cdaf}
A \emph{context-dependent argumentation framework} (CDAF) is a tuple $D = \langle \Args, R, \Contexts, \delta \rangle$, where $\Args$ is a finite set of arguments, $R \subseteq \Args \times \Args$ is an attack relation, $\Contexts$ is a non-empty finite set of contexts, and $\delta\colon \Contexts \times R \to \{0,1\}$ is the \emph{defeat function}.
\end{definition}

Each context $c \in \Contexts$ induces the defeat relation $R_c = \{(a,b) \in R : \delta(c,(a,b)) = 1\}$ and the Dung framework $AF_c = \langle \Args, R_c \rangle$; a suppressed attack is deleted, not reversed (see Section \ref{sec:representability}). Attacks in $R$ are \emph{latent} conflicts; attacks in $R_c$ are the \emph{operative} defeats in $c$. We write $k = |\Contexts|$.

Cross-context acceptance classifies arguments across the family. Under a semantics $\sigma$, an argument is \emph{universally accepted} if it belongs to every $\sigma$-extension of every $AF_c$, \emph{indefensible} if it belongs to none, and \emph{context-contingent} otherwise.

\section{Representability and Its Algorithm}\label{sec:representability}

\textsc{VAF-Representability} asks whether audiences suffice to explain a CDAF's variation. Given $D = \langle \Args, R, \Contexts, \delta \rangle$, are there a value set $V$, an assignment $\mathit{val}\colon \Args \to V$, and a map $\varphi$ from $\Contexts$ to strict total orders on $V$ with $\delta(c,(a,b)) = 1$ iff $\mathit{val}(b) \not\succ_{\varphi(c)} \mathit{val}(a)$, for every $c \in \Contexts$ and $(a,b) \in R$? The VAF must reproduce every entry of $\delta$; matching only extensions is the weaker realisability problem of Section \ref{sec:complexity}.

\subsection{The criterion and the algorithm}

The criterion is a partition condition. For a partition $\mathcal{P}$ of $\Args$, write $[a]$ for the class of argument $a$.

\begin{definition}[value-consistent partition]\label{def:vcp}
A partition $\mathcal{P}$ of $\Args$ is \emph{value-consistent} for $D$ if the following hold.
\begin{itemize}
\item[(I)] \emph{Intra-class unanimity:} for all $(a,b) \in R$ with $[a] = [b]$ and all $c \in \Contexts$: $\delta(c,(a,b)) = 1$.
\item[(U)] \emph{Cross-class uniformity:} for all $(a_1,b_1), (a_2,b_2) \in R$ with $a_1, a_2 \in [\alpha]$, $b_1, b_2 \in [\beta]$, and $[\alpha] \neq [\beta]$: $\delta(c,(a_1,b_1)) = \delta(c,(a_2,b_2))$ for all $c \in \Contexts$.
\item[(C)] \emph{Cross-class complementarity:} for all $(a,b), (b',a') \in R$ with $a, a' \in [\alpha]$, $b, b' \in [\beta]$, and $[\alpha] \neq [\beta]$: $\delta(c,(a,b)) + \delta(c,(b',a')) = 1$ for all $c \in \Contexts$.
\item[(A)] \emph{Cross-class acyclicity:} for each $c \in \Contexts$ the relation $\rhd_c$ on $\mathcal{P}$ is acyclic, where for distinct classes $[\alpha] \rhd_c [\beta]$ iff some attack from $[\alpha]$ to $[\beta]$ succeeds in $c$, or some attack from $[\beta]$ to $[\alpha]$ fails in $c$.
\end{itemize}
\end{definition}

\begin{theorem}\label{thm:criterion}
A CDAF is VAF-representable iff it admits a value-consistent partition.
\end{theorem}

The conditions unfold the arithmetic of strict total orders. Equal values never block a defeat, so intra-class attacks must always succeed, which is (I). An audience compares values rather than individual attacks, so parallel cross-class attacks share one outcome, which is (U), and opposite cross-class attacks have exactly one winner, which is (C). A successful attack from $[\alpha]$ to $[\beta]$ and a failed attack from $[\beta]$ to $[\alpha]$ both witness that $[\alpha]$'s value is strictly preferred in $c$; $\rhd_c$ collects these forced preferences, and by Szpilrajn's theorem \cite{szpilrajn1930} they extend to a strict total order iff they contain no cycle, which is (A). Conversely, take $V = \mathcal{P}$, $\mathit{val}(a) = [a]$, and any $\succ_c$ extending $\rhd_c$. Under (U) and (C) the two clauses never conflict, so $\rhd_c$ is well defined and antisymmetric, and three cases reproduce $\delta$: an intra-class attack has equal values, so it defeats, and (I) gives $\delta = 1$; a succeeding cross-class attack gives $[a] \rhd_c [b]$, hence $\mathit{val}(b) \not\succ_c \mathit{val}(a)$; a failing one gives $[b] \rhd_c [a]$, hence $\mathit{val}(b) \succ_c \mathit{val}(a)$.

The search for a partition is polynomial because all merges are forced: Algorithm \ref{alg:rep} starts from singleton classes and only ever merges.

\begin{algorithm}[h]
\caption{Deciding VAF-representability of a CDAF $\langle \Args, R, \Contexts, \delta \rangle$}
\label{alg:rep}
\begin{algorithmic}[1]
\State $\mathcal{P} \gets \{\{a\} : a \in \Args\}$ \Comment{union-find over $\Args$}
\Statex \emph{Phase 1: co-equality merges}
\ForAll{mutual pairs $(a,b), (b,a) \in R$}
  \If{$\delta(c,(a,b)) = 1$ and $\delta(c,(b,a)) = 1$ for every $c \in \Contexts$}
    \State merge the classes of $a$ and $b$
  \EndIf
\EndFor
\Repeat
  \Statex \quad \emph{Phase 2: intra-class check (the only failure point)}
  \ForAll{$(a,b) \in R$ with $[a] = [b]$}
    \If{$\delta(c,(a,b)) = 0$ for some $c \in \Contexts$}
      \State \Return not representable
    \EndIf
  \EndFor
  \Statex \quad \emph{Phase 3: per-context cycle merges}
  \State $\mathit{merged} \gets \textbf{false}$
  \ForAll{$c \in \Contexts$}
    \State build $\rhd_c$ on the classes of $\mathcal{P}$; record its non-trivial SCCs
  \EndFor
  \ForAll{recorded non-trivial SCCs $S$}
    \State merge the classes in $S$; $\mathit{merged} \gets \textbf{true}$
  \EndFor
\Until{$\mathit{merged} = \textbf{false}$}
\Statex \emph{Phase 4: verification}
\State check (U), (C), (A) for $\mathcal{P}$
\State \Return $\mathcal{P}$ \Comment{the finest value-consistent partition}
\end{algorithmic}
\end{algorithm}

\begin{theorem}\label{thm:poly}
VAF-representability is decidable in $O(n^2 k + nmk)$ time.
\end{theorem}

The correctness argument rests on two invariants. First, at every stage the current partition refines every value-consistent partition, because Phase 1 merges are forced by (C) and Phase 3 merges by (A): a $\rhd_c$ cycle over distinct classes can only be repaired by making them coincide. Second, any partition reaching Phase 4 is value-consistent, and this is checkable rather than asserted: between two distinct classes, a violation of (U) or (C) in a context $c$ always produces a two-cycle in $\rhd_c$. For (U), the succeeding parallel attack gives $[\alpha] \rhd_c [\beta]$ and the failing one gives $[\beta] \rhd_c [\alpha]$; for (C), both succeeding and both failing each give edges in both directions. Phase 3 merges every such cycle, so Phase 4 never fails; Phase 1 is an optimisation, its merges being (C) violations a Phase 3 pass would find anyway. The algorithm therefore rejects exactly the non-representable instances and otherwise returns the unique finest value-consistent partition. Phase 3 tests each $\rhd_c$ separately, never their union: condition (A) is imposed per context, and edges from different contexts may form a union cycle while every single $\rhd_c$ stays acyclic.

For the running time: each round costs $O(mk)$ to run Phase 2 and build the relations $\rhd_c$. Each $\rhd_c$ carries at most $m$ edges, one per attack, so its strongly connected components cost $O(n+m)$, or $O((n+m)k)$ per round. Every merging round loses at least one class, so there are at most $n-1$ of them, plus one final round.

\paragraph{Blocked attacks: deletion or inversion.} We read a suppressed attack as deleted. The companion paper \cite{sadowski2026choosinglensstrategicperspective} reads it as inverted, in the style of the critical-attack repair of \cite{amgoud2014}. Deletion is the reading under which the VAF question has content: an audience's defeats always lie inside $R$, whereas inversion can produce a defeat outside $R$, and does so exactly when a one-directional attack is blocked. Representability under inversion therefore requires (S): $\delta(c,(a,b)) = 1$ whenever $(a,b) \in R$ and $(b,a) \notin R$. Given (S) the two readings agree except on mutual pairs blocked in both directions, where inversion supplies both edges and deletion neither. Flipping exactly those pairs to active gives a defeat function $\delta^*$, computable in $O(mk)$, and a CDAF is representable under inversion iff (S) holds and $\langle \Args, R, \Contexts, \delta^* \rangle$ is representable under deletion. Theorem \ref{thm:criterion} and Algorithm \ref{alg:rep} therefore transfer after one linear-time pass, with Theorem \ref{thm:poly} unchanged. Condition (S) is itself a separation: under inversion, VAF-representability forces all blocking onto mutual attacks.

\subsection{A worked trace}\label{sec:trace}

We trace Algorithm \ref{alg:rep} on an instance built so that one run exercises every phase: a Phase 1 merge, a Phase 3 cycle merge, and a Phase 2 rejection.

Let $\Args = \{w, x, y, z\}$ and $\Contexts = \{c_1, c_2\}$, with
\[
R = \{(w,x),\ (x,w),\ (x,y),\ (y,z),\ (z,x)\}.
\]
The graph is a mutual pair $\{w,x\}$ plus a directed cycle $x \to y \to z \to x$; $c_1$ activates every attack and $c_2$ suppresses only $(x,y)$, as Figure \ref{fig:instance} shows.

\begin{figure}[t]
\centering
\begin{minipage}[c]{0.48\linewidth}
\centering
\begin{tikzpicture}[arg/.style={draw, circle, inner sep=1.5pt, minimum size=5.5mm}]
\node[arg] (w) at (0,0) {$w$};
\node[arg] (x) at (1.8,0) {$x$};
\node[arg] (y) at (3.6,0) {$y$};
\node[arg] (z) at (2.7,-1.5) {$z$};
\draw[->] (w) to[bend left=18] (x);
\draw[->] (x) to[bend left=18] (w);
\draw[->, dashed] (x) -- (y);
\draw[->] (y) -- (z);
\draw[->] (z) -- (x);
\end{tikzpicture}
\end{minipage}\hfill
\begin{minipage}[c]{0.48\linewidth}
\centering
\begin{tabular}{lcc}
\toprule
Attack & $c_1$ & $c_2$ \\
\midrule
$(w,x)$ & 1 & 1 \\
$(x,w)$ & 1 & 1 \\
$(x,y)$ & 1 & 0 \\
$(y,z)$ & 1 & 1 \\
$(z,x)$ & 1 & 1 \\
\bottomrule
\end{tabular}
\end{minipage}
\caption{The worked instance. Left: the attack relation $R$; the dashed attack $(x,y)$ is the only attack suppressed in some context. Right: the defeat function $\delta$.}
\label{fig:instance}
\end{figure}

\begin{enumerate}
\item \emph{Phase 1.} The only mutual pair is $\{w, x\}$. Both $(w,x)$ and $(x,w)$ succeed in both contexts, so $w$ and $x$ merge. Now $\mathcal{P} = \{\{w,x\}, \{y\}, \{z\}\}$.
\item \emph{Phase 2.} The intra-class attacks are $(w,x)$ and $(x,w)$, and both succeed in every context. Pass.
\item \emph{Phase 3, building the relations.} On the classes $[wx]$, $[y]$, $[z]$. In $c_1$ every attack succeeds, so the first clause yields $[wx] \rhd_{c_1} [y]$, $[y] \rhd_{c_1} [z]$, and $[z] \rhd_{c_1} [wx]$: a cycle, one non-trivial strongly connected component. In $c_2$ the attack $(x,y)$ fails, so the second clause yields $[y] \rhd_{c_2} [wx]$; with $[y] \rhd_{c_2} [z]$ and $[z] \rhd_{c_2} [wx]$ from the first clause, $\rhd_{c_2}$ is acyclic. So $c_2$ forces preferences but no merge; the cycle lives in $\rhd_{c_1}$ alone.
\item \emph{Phase 3, merging.} The $c_1$ component merges all three classes: $\mathcal{P} = \{\{w,x,y,z\}\}$. Back to Phase 2.
\item \emph{Phase 2, second pass.} Every attack is now intra-class. The check reaches $(x,y)$ and finds $\delta(c_2,(x,y)) = 0$. The algorithm returns \emph{not representable}, at its only failure point.
\end{enumerate}

The rejection tracks a semantic impossibility that can be stated without the algorithm. Suppose a value-consistent partition existed. In $c_1$ all three attacks of the cycle $x \to y \to z \to x$ succeed. If $x$, $y$, $z$ did not all share one class, the distinct classes among them would carry a $\rhd_{c_1}$ cycle, violating (A). So $[x] = [y] = [z]$. But then $(x,y)$ is an intra-class attack, and $\delta(c_2,(x,y)) = 0$ violates (I).

Neither context is pathological alone. Each is representable in isolation: $c_1$ by one value for every argument, $c_2$ by $\mathit{val}(w) = \mathit{val}(x) = v$ with $y$ and $z$ given fresh values ordered $\mathit{val}(y) \succ \mathit{val}(z) \succ v$. Non-representability is a coupling phenomenon: it appears only when one value assignment must serve both contexts.

\section{The Complexity Landscape}\label{sec:complexity}

Table \ref{tab:landscape} collects the decision problems that a CDAF generates and what is known about them.

\begin{table}[t]
\caption{Decision problems for a CDAF with $n$ arguments, $m$ attacks, and $k$ contexts. Cross-context bounds follow from single-framework classifications \cite{dvorak2018complexity} with a factor $k$ for iterating over contexts; the corresponding audience-dependent questions for VAFs are classified by \cite{BENCHCAPON200742}.}
\label{tab:landscape}
\centering
\begin{tabular}{llll}
\toprule
Problem & Semantics & Bound & Status \\
\midrule
Cross-context acceptance (all classes) & grounded & in P & closed \\
Universal acceptance & preferred & $\Pi^p_2$-complete & closed \\
Indefensibility & preferred & coNP-complete & closed \\
Audience inducibility of $R' \subseteq R$ & (defeat relation) & $O(n+m)$ & closed \\
VAF-representability & (defeat function) & $O(n^2 k + nmk)$ & closed \\
Single-context realisability & grounded & in NP & lower bound open \\
Single-context realisability & preferred & in $\Sigma^p_3$ & lower bound open \\
Family realisability & any & decidable & criterion open \\
\bottomrule
\end{tabular}
\end{table}

\paragraph{Single-audience inducibility.} The basic building block, reused by the implementation of Section \ref{sec:benchmarks}, is a linear-time test.

\begin{proposition}[mixed-graph inducibility test]\label{prop:induce}
Given $\langle \Args, R \rangle$ and $R' \subseteq R$, deciding whether $R' = R_p$ for some audience $p$ of some VAF over $\langle \Args, R \rangle$ takes $O(n+m)$ time.
\end{proposition}

Build a mixed graph $H$ on the arguments: a weak edge $b \to a$ for each $(a,b) \in R'$, and a strict edge $a \to b$ for each $(a,b) \in R \setminus R'$. Read an edge $u \to v$ as ``the rank of $u$ is at most the rank of $v$'', strictly so for strict edges. The constraint system is satisfiable iff $H$ has no directed cycle through a strict edge: contract the strongly connected components and reject iff some strict edge lies inside one.

\paragraph{Single-context realisability.} Realisability weakens representability: the VAF only has to reproduce the extensions. For a context $c$ and semantics $\sigma$, the question is whether some audience $p$ satisfies $\sigma(\langle \Args, R_p \rangle) = \sigma(AF_c)$. Guess and check gives the upper bounds of Table \ref{tab:landscape}, neither known to be tight. The natural criterion for the grounded case fails: acyclicity of the latent attacks inside the target extension is necessary but not sufficient, and the smallest counterexamples, found by exhaustive enumeration up to four arguments, have three arguments and four attacks. Grounded and preferred realisability also come apart: some instances are realisable under grounded but not under preferred semantics.

The strategic variant is the subject of \cite{sadowski2026choosinglensstrategicperspective}: choosing which perspectives are active selects which attacks succeed, and deciding whether some choice makes a target argument credulously accepted is NP-complete under preferred and stable semantics.

\section{A Reference Implementation and a Measurement}\label{sec:benchmarks}

We implemented Algorithm \ref{alg:rep} and the inducibility test of Proposition \ref{prop:induce} as a Python script.\footnote{We release the implementation, the validation suite, and the enumeration data at \url{https://github.com/albsadowski/cdaf}, archived at \url{https://doi.org/10.5281/zenodo.21320967}. Small complete instance spaces support conjecture testing, as in the realisability and signature programme \cite{dunne2015characteristics,linsbichler2016uniform}; the grounded counterexample of Section \ref{sec:complexity} came from exactly such an enumeration.} The tool reads a small extension of the apx format used in ICCMA \cite{thimm2017first,gaggl2020design,bistarelli2025iccma,jarvisalo2025iccma}: a context block declares the contexts and lists, per context, the suppressed attacks. The worked instance of Section \ref{sec:trace} reads:

\begin{lstlisting}[basicstyle=\ttfamily\small]
arg(w). arg(x). arg(y). arg(z).
att(w,x). att(x,w). att(x,y). att(y,z). att(z,x).
context(c1).
context(c2). off(c2,x,y).
\end{lstlisting}

Every attack is operative unless switched off, so a plain apx file is a CDAF with one context, and projecting to one context yields a plain apx instance on which any ICCMA solver runs unchanged; cross-context queries are thin orchestration over $k$ solver calls.

Correctness is machine-checked against the definitions. One brute-force oracle enumerates all partitions of $\Args$ and checks the four conditions of Definition \ref{def:vcp} literally; a second decides inducibility by trying every ranking. The implementation agrees with the oracles on all 547,795 instances with three arguments and up to three contexts, on 180,000 uniformly sampled instances with four and five arguments, and on all 532,180 inducibility pairs with up to four arguments. A third oracle decides representability directly from the defeat rule, trying every value assignment and every per-context strict total order; it agrees on all 547,795 three-argument instances, and since it does not use the criterion, that agreement checks Theorem \ref{thm:criterion} itself, not only the algorithm. The returned partition is checked value-consistent and finest on every representable instance, and the worked trace is asserted event by event.

On VAF-generated instances, representable by construction and so necessarily accepted, the reference code handles $10^3$ arguments, $10^5$ attacks, and 20 contexts in five seconds; a mask-based variant in the release handles $10^4$ arguments, $2 \times 10^6$ attacks, and 20 contexts in under three minutes.

\subsection{How common is representability under uniform sampling?}

The same machinery yields a measurement: the fraction of small CDAFs that is VAF-representable. Table \ref{tab:fractions} gives it for labeled instances without self-attacks. The generator is uniform over that space: each ordered pair of distinct arguments independently takes one of $1 + 2^k$ equiprobable states, either absent from $R$ or present with one of the $2^k$ activation patterns over the contexts. So $R$ and $\delta$ are drawn in one step, not independently, and every attack is equally likely to carry any pattern. This is a null model, not a model of applications.

\begin{table}[t]
\caption{Percentage of VAF-representable CDAFs among labeled instances without self-attacks, by arguments $n$ and contexts $k$. The cells with $n=3$, $k \le 3$ are exact (exhaustive enumeration); the others are estimates from $2 \times 10^5$ uniform samples per cell, all 95\% confidence intervals below $0.2$ percentage points. Cells with no representable sample are bounded above by $0.002$ percent.}
\label{tab:fractions}
\centering
\begin{tabular}{lcccc}
\toprule
 & $k=1$ & $k=2$ & $k=3$ & $k=4$ \\
\midrule
$n=3$ & $54.9$ & $10.5$ & $1.51$ & $0.18$ \\
$n=4$ & $20.3$ & $0.52$ & $0.009$ & $<0.002$ \\
$n=5$ & $4.1$ & $0.003$ & $<0.002$ & $<0.002$ \\
\bottomrule
\end{tabular}
\end{table}

Representability collapses in both directions, from 55 percent at $n=3$, $k=1$ to under one instance in ten thousand at $n=5$, $k=2$. Beyond $n=5$ it falls under the detection limit of uniform sampling, so larger instances would add empty bounds, not measurements. Two readings. First, audience structure is a strong constraint: generic context-dependence is almost never reducible to value orderings. The scope of that claim is the null model above; structured instances from applications, such as attack graphs extracted from text by argument-mining pipelines \cite{baba2026argumentcomponentsgraphsmultiagent}, are not covered by these numbers. Second, instances satisfying three of the four conditions of Definition \ref{def:vcp} are too rare to sample and must be constructed; the enumeration data locates them. Realisability labels, by contrast, are not free at generation time: they need extension-set comparisons, at the second level of the polynomial hierarchy for preferred semantics, which current competition tracks do not exercise.

\section{Open Problems}\label{sec:open}

The main structural question is family-level realisability, where one value assignment couples all contexts: a criterion in the style of Theorem \ref{thm:criterion}, or any upper bound better than brute force. For the grounded single-context case, acyclicity of latent attacks inside the target extension is necessary but not sufficient, and a characterisation would settle whether the NP bound can be lowered; more broadly, the realisability upper bounds have no matching lower bounds. Whether the perspective-labeled fragment of \cite{sadowski2026choosinglensstrategicperspective}, which derives $\delta$ from perspective activation and priorities, admits faster algorithms is unexplored. A separate question is where contexts come from: they are external parameters here, whereas dialogue supplies them endogenously, a protocol stage or a shift in the burden of proof being a natural generator of $\delta$. Whether dialogue protocols characterise the CDAF families that arise this way is open. Finally, cross-context queries materialise $k$ frameworks and run a solver per context; declarative encodings working on $\delta$ directly, in the spirit of the ASPIC+ encodings of \citet{lehtonen2024Complexity}, would avoid that and are a natural next step for this community.

\section{Conclusion}\label{sec:conclusion}

A CDAF is one attack structure under many contexts. VAF-representability is solved by a polynomial-time partition-refinement algorithm, validated against the definitions on over half a million small instances, while the surrounding realisability problems have upper bounds from NP to $\Sigma^p_3$ and no matching lower bounds. The measurement shows where the algorithmic interest lies: under uniform sampling representability is thin and thins with every added context, so cross-context reasoning cannot in general be delegated to VAF machinery. The open problems of Section \ref{sec:open} mark where new criteria and encodings are needed; the released code and data are a starting point.

\bibliographystyle{unsrtnat}
\bibliography{references}

\end{document}